\pdfoutput=1

\documentclass[11pt]{article}
\usepackage{xcolor}
\usepackage{tabularx}
\usepackage{float}
\usepackage{EMNLP2023}
\usepackage{amsmath} 
\usepackage{booktabs}
\usepackage{indentfirst}
\usepackage{times}
\usepackage{latexsym}
\usepackage{graphicx}
\usepackage[T1]{fontenc}

\usepackage[utf8]{inputenc}
\usepackage{textcomp}
\usepackage{needspace}
\usepackage{enumitem}
\usepackage{tcolorbox}
\tcbuselibrary{breakable}

\usepackage{microtype}

\usepackage{inconsolata}
\newcommand{\ftmodel}[1]{\textcolor{blue}{#1}}

\tcbset{
  colback=gray!4,
  colframe=gray!65,
  boxrule=0.5pt,
  arc=2pt,
  width=\linewidth,
  breakable,
  fonttitle=\bfseries,
  left=4pt,
  right=4pt,
  top=3pt,
  bottom=3pt
}
\title{OntologyBench: Can Dense Retrieval Satisfy Structured Biomedical Constraints?}

\author{
Xiao Yu Cindy Zhang$^{1,2}$ \quad
Wyeth W. Wasserman$^{2,3}$ \quad
Jian Zhu$^{4}$ \\[0.4em]
$^{1}$Graduate Program in Bioinformatics, The University of British Columbia \\
$^{2}$Centre for Molecular Medicine and Therapeutics, BC Children's Hospital Research Institute \\
$^{3}$Department of Medical Genetics, The University of British Columbia \\
$^{4}$Department of Linguistics, The University of British Columbia \\
Vancouver, BC, Canada \\[0.3em]
\texttt{czhang@cmmt.ubc.ca} \quad
\texttt{wyeth@cmmt.ubc.ca} \quad
\texttt{jian.zhu@ubc.ca}
}

\begin{document}
\maketitle
\begin{abstract}
We introduce OntologyBench, a tiered biomedical retrieval benchmark comprising 471,854 training and 125,744 evaluation query–document relevance pairs across concept grounding, relational retrieval, and compositional phenotype-based retrieval.

Although these tasks can be tractable using ontology-aware reference methods, across task tiers, embedding performance is generally lower on relational and compositional tasks than on concept-grounding tasks. Fine-tuning on ontology-derived supervision improves performance on several relational and compositional tasks, whereas the evaluated reranking and LLM-based candidate-scoring methods provide little or no end-to-end improvement. Errors frequently reflect diseases matching only subsets of the phenotype evidence. 

These findings indicate that the evaluated embedding and reranking configurations do not reliably recover the compatibility encoded by the selected ontology relations and phenotype combinations and motivate retrieval systems that better integrate learned representations with structured biomedical knowledge.

\end{abstract}

\section{Introduction}

Dense retrieval models enable retrieval directly over natural language by mapping semantically related queries and documents into a shared embedding space. This supports retrieval beyond exact lexical overlap and reduces dependence on explicitly engineered symbolic methods, such as ontology-based semantic similarity scoring and graph traversal. This flexibility has made embedding retrieval attractive for biomedical and clinical applications, where concepts are often expressed using heterogeneous, incomplete, or non-canonical language.

However, many retrieval problems require satisfying multiple interdependent constraints rather than matching individual features independently. \citep{agarwal2026rear} Existing benchmarks primarily evaluate retrieval based on unstructured semantic similarity and provide limited insight into whether models can recover compatibility across multiple structured inputs. \citep{muennighoff2023mteb}

To address this gap, we introduce OntologyBench, a retrieval benchmark constructed from expert-curated biomedical ontologies. The benchmark spans concept grounding, relational retrieval, and compositional phenotype-based retrieval. These task families provide complementary views of retrieval behaviour under different forms of ontology-encoded compatibility. Because they also differ in candidate spaces, relevance structures, and query ambiguity, cross-tier comparisons are interpreted descriptively rather than as isolated estimates of constraint complexity.

Within this framework, the fixed Tier~3 phenotype-triplet-to-disease task reveals a persistent limitation. The evaluated embedding models frequently rank diseases that match only a subset of the phenotype evidence, and their performance remains substantially below that of an ontology-aware baseline. The tested rerankers provide little or no end-to-end improvement across the complete query set. These findings indicate that the evaluated retrieval systems do not reliably recover the compatibility encoded by joint phenotype sets, although candidate omission and downstream ranking errors remain alternative sources of failure.

This work makes three contributions. First, we introduce a retrieval benchmark spanning concept grounding, ontology-relation retrieval, and phenotype-set retrieval to evaluate whether embedding models recover compatibility encoded by curated biomedical ontologies. Second, we characterize retrieval behaviour across grounding, relational, and compositional tasks, architectures, and model scales, identifying a persistent gap relative to an ontology-aware baseline on the fixed phenotype-triplet-to-disease task. Third, through two-stage evaluation, we distinguish candidate availability from the end-to-end performance of the reranked system. Across the complete query set, the evaluated rerankers provide little or no improvement over the first-stage retriever.

\section{Related Work}

\paragraph{Dense Retrieval Architectures.}
Dense retrieval represents queries and documents as vectors in a shared embedding space, enabling efficient large-scale search~\citep{reimers2019sentence}. Bi-encoder models support scalable retrieval via independent encoding, while late-interaction models such as ColBERT preserve token-level representations to improve fine-grained matching~\citep{khattab2020colbert, chaffin2025pylate}. Recent pretrained models further improve embedding quality for retrieval tasks~\citep{zhang2025qwen3}.  

\paragraph{Embedding Benchmarks.}
Large-scale benchmarks such as BEIR~\citep{thakur2021beir} and MTEB~\citep{muennighoff2023mteb} standardize evaluation across diverse retrieval and embedding tasks, primarily measuring semantic similarity in unstructured text. Biomedical retrieval datasets such as NFCorpus~\citep{boteva2016full} and TREC-COVID~\citep{voorhees2021trec} similarly evaluate retrieval between clinical documents. More recent benchmarks explore retrieval requiring aggregation across multiple documents or reasoning steps, including multi-hop question answering~\citep{yang2018hotpotqa} and reasoning-intensive retrieval benchmarks~\citep{xiao2024rar, su2024bright, li2025r2med}. However, these benchmarks operate over unstructured text and do not isolate compositional compatibility.

Biomedical retrieval has long relied on symbolic methods that model relationships between diseases, phenotypes, and genes. Information-content–based semantic similarity methods, including Resnik similarity and phenotype set similarity approaches used in systems such as Phenomizer, leverage hierarchical biomedical structure for retrieval. \citep{resnik1999semantic, kohler2009clinical, hoehndorf2015role, vasilevsky2025mondo} Graph-based retrieval methods similarly exploit relational paths between biomedical entities. Unlike embedding-based retrieval, these approaches enforce compatibility through curated symbolic structure rather than learning it implicitly through representation similarity, making them less flexible for heterogeneous natural-language inputs such as free-text clinical documentation.

Recent work has also explored ontology-grounded biomedical embeddings designed to improve interpretability. QIME~\citep{tang2026qime} constructs interpretable medical text embeddings by representing documents through ontology-grounded natural-language questions derived from biomedical concepts. While such approaches improve semantic interpretability and retrieval effectiveness, they primarily evaluate embedding quality through semantic similarity, clustering, and retrieval performance rather than through controlled evaluation of compositional compatibility. In contrast, OntologyBench focuses on whether retrieval systems can jointly satisfy multiple structured biomedical constraints under controlled retrieval settings.

OntologyBench evaluates whether embedding-based retrieval systems can recover the compatibility constraints enforced by symbolic biomedical retrieval methods. While embedding retrieval supports flexible natural-language matching, symbolic biomedical systems directly enforce compatibility through curated relational structure. OntologyBench  organizes retrieval tasks across concept grounding, relational retrieval, and phenotype-set retrieval, enabling comparison of retrieval behaviour under different forms of ontology-defined relevance.

\section{OntologyBench Dataset and Tasks}

\begin{figure*}[t]
    \centering
    \includegraphics[width=\textwidth, height=9cm]{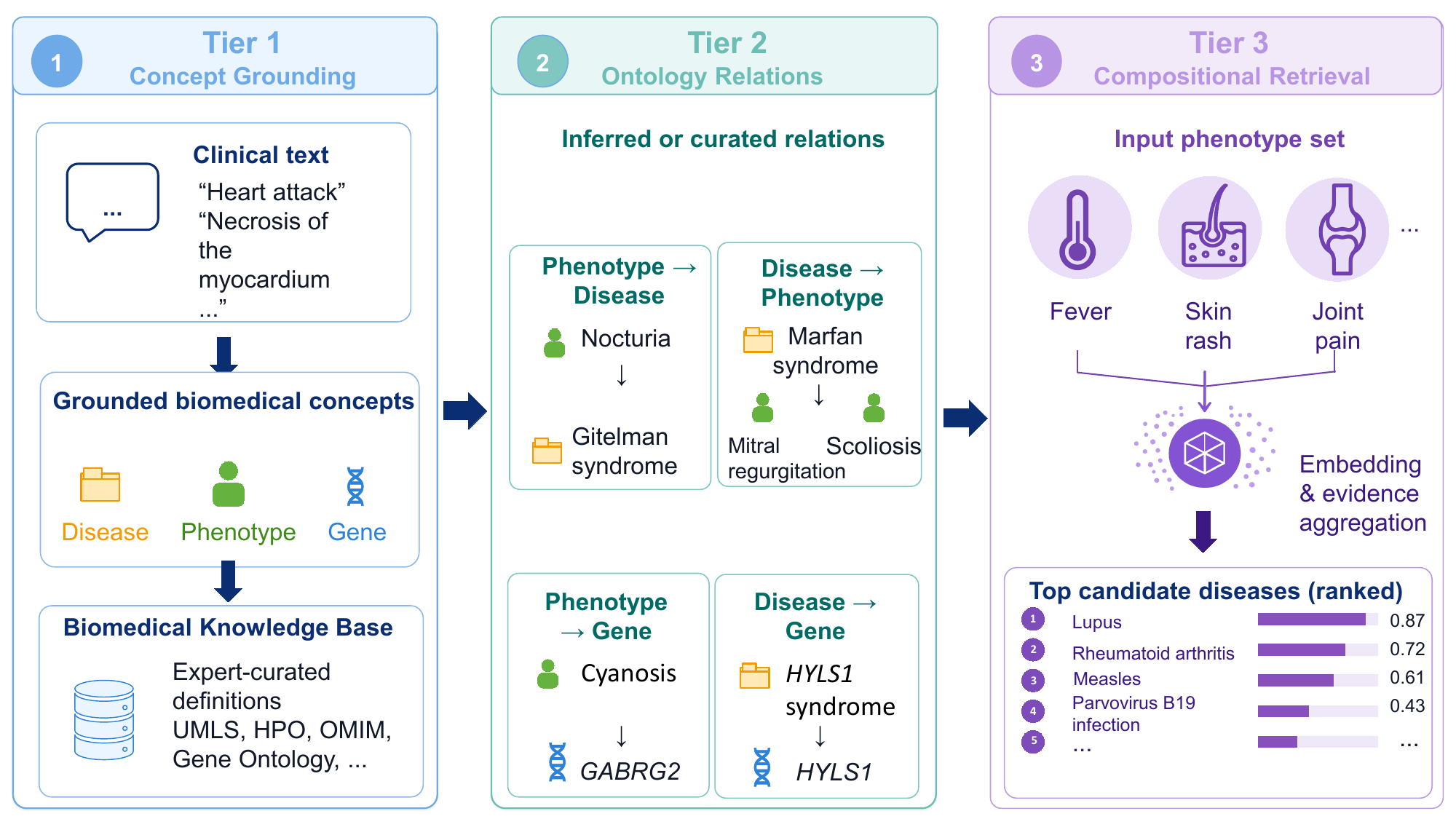}
    \caption{
OntologyBench overview illustrating the retrieval tiers from grounding to compositional retrieval. 
}
    \label{fig:ontologybench}
\end{figure*}

\subsection{Knowledge Sources.}

OntologyBench is designed as an evaluation framework spanning concept grounding, ontology-relation retrieval, and phenotype-set retrieval. The task families differ in candidate spaces, relevance structures, and query ambiguity; cross-tier comparisons are therefore interpreted descriptively. The unified setting evaluates held-out directional pairs and phenotype combinations within a shared ontology vocabulary rather than generalization to entirely unseen biomedical concepts or relations.

We construct a structured retrieval space grounded in curated biomedical ontologies linking clinical findings, diseases, and genetic causes. Disease concepts are drawn from the Monarch Disease Ontology (MONDO) \citep{vasilevsky2025mondo}, which integrates disease terminology and mappings across resources including OMIM, SNOMED CT, and ICD. OntologyBench includes 6,611 canonical MONDO disease concepts represented by their preferred labels and available synonyms.

Phenotype concepts are drawn from the Human Phenotype Ontology (HPO)~\citep{gargano2024mode}, a controlled vocabulary describing observable clinical features (phenotypes). The dataset contains 8,253 phenotype concepts, each with associated textual synonyms used as query inputs.

Gene entities are constructed from HUGO Gene Nomenclature Committee(HGNC) gene names~\citep{seal2026genenames} combined with National Center for Biotechnology Information
 (NCBI) Gene metadata~\citep{o2024exploring}, resulting in 2,345 gene entries.

Together, these ontologies provide a structured biomedical knowledge space linking phenotypes, diseases, and genes, enabling evaluation of retrieval systems under structured biomedical constraints.

\subsection{Task Formulation}
OntologyBench is organized into three tiers spanning concept grounding, ontology relations, and compositional retrieval (Figure~\ref{fig:ontologybench}). This organization supports comparison across concept grounding, ontology-relation retrieval, and phenotype-set retrieval. Because the task families differ in candidate spaces, relevance multiplicity, and query ambiguity, cross-tier score differences are descriptive rather than isolated effects of increasing constraint complexity. Examples of each tier can be found in Appendix~\ref{appendix:examples}.

\paragraph{Tier~1: Concept Grounding}
Tier~1 evaluates whether models can map surface forms (aliases) of genes (Gene), phenotypes (Phen), and disorders (Dis) to their canonical ontology definitions (Def) through single-hop retrieval. This tier includes three grounding tasks: \textbf{Gene→Def}, \textbf{Phen→Def}, and \textbf{Dis→Def}.

\paragraph{Tier~2: Relational Retrieval.}

This tier evaluates whether models capture pairwise biomedical relationships between diseases, phenotypes, and genes.

\textbf{Disease $\leftrightarrow$ Phenotype (Dis$\leftrightarrow$Phen).}
These tasks evaluate complementary retrieval patterns over the same curated
many-to-many disease--phenotype relation. Disease-to-phenotype retrieval
identifies characteristic features for constructing or reviewing disease
profiles, whereas phenotype-to-disease retrieval identifies diseases
compatible with an observed finding and is generally more ambiguous because
phenotypes are shared across diseases. Their differing query semantics,
candidate spaces, and relevance-set structures motivate evaluation in both
directions.

\textbf{Phenotype→Gene (Phen$\rightarrow$Gene).}
Given a phenotype query, the model retrieves genes associated with diseases exhibiting that phenotype.

\textbf{Disease→Gene (Dis$\rightarrow$Gene).}
Given a disease query, the model retrieves genes known to be molecularly associated with that disorder.

\paragraph{Tier~3: PhenTriplet→Dis (compositional retrieval).}

In clinical settings, diagnosis is often based on incomplete and partially observed phenotypes. Tier~3 models this under a controlled phenotype-based retrieval setting, evaluating compositional retrieval from limited evidence. Each query consists of three phenotypes drawn from a single disease’s phenotype set, ensuring internal consistency. Some phenotype subsets may include non-specific features; this is intentional, as such cases introduce ambiguity and test whether models can correctly aggregate weak signals rather than rely on a single highly discriminative feature.

The choice of three phenotypes reflects a trade-off between ambiguity and determinism: single phenotypes correspond to many candidate diseases, while larger sets ($\geq4$ phenotypes) sharply reduce the candidate space, making retrieval increasingly deterministic. As shown in Appendix~\ref{tier3:cardinality}, phenotype triplets yield a non-trivial regime ($\sim5$ relevant diseases on average), requiring models to aggregate multiple constraints rather than rely on single-feature matching.

While this construction does not model the full diagnostic process, which often needs to incorporate temporal progression, demographics, disease prevalence and other contextual evidence, it intentionally isolates the compositional retrieval problem by controlling for these additional factors. This enables evaluation of whether retrieval models can integrate multiple phenotype constraints under ambiguity.  

\begin{table}[t]
\centering
\small
\setlength{\tabcolsep}{4pt}
\renewcommand{\arraystretch}{0.5}
\begin{tabular}{lcc|cc}
\toprule
& \multicolumn{2}{c}{Train} & \multicolumn{2}{c}{Test} \\
\cmidrule(lr){2-3} \cmidrule(lr){4-5}
Task & \#Queries & AvgPos & \#Queries & AvgPos \\
\midrule

\multicolumn{5}{l}{\textbf{Tier~1 (Concept Grounding)}} \\

Gene→Def & 1,914 & 1.00 & 431 & 1.00 \\
Phen→Def & 17,763 & 1.00 & 4,581 & 1.00 \\
Dis→Def & 39,644 & 1.00 & 9,205 & 1.00 \\

\midrule
\multicolumn{5}{l}{\textbf{Tier~2 (Relational Retrieval)}} \\

Dis→Phen & 6,647 & 19.81 & 6,291 & 5.63 \\
Phen→Dis & 7,988 & 16.62 & 5,229 & 6.57 \\

Phen→Gene & 6,960 & 1.00 & 1,507 & 1.00 \\
Dis→Gene & 6,623 & 17.58 & 6,504 & 5.31 \\

\midrule
\multicolumn{5}{l}{\textbf{Tier~3 (Compositional Retrieval)}} \\

PhenTriplet→Dis & 5,642 & 4.39 & 2,537 & 2.26 \\

\bottomrule
\end{tabular}

\caption{
OntologyBench dataset statistics across tiers. AvgPos denotes the average number of positive documents per query. Tier-3 queries had an average of 4.39 relevant diseases in training and 2.26 in evaluation.
}
\label{tab:ontologybench_stats}
\end{table}

\subsection{Train–Evaluation Partition and Split-Integrity Audit}

The unified multi-tier corpus comprises 471,854 training and 125,744
evaluation query--document relevance pairs.  Task-specific partitions were
constructed at the level of canonical directed query--target identifiers, and
an audit of the training artifacts found zero within-task overlap of these
directed pairs between training and evaluation. Because the tasks share an
ontology vocabulary, concepts, targets, and underlying associations may recur
across task formulations.

The two directions of Tier~2 Dis$\leftrightarrow$Phen were partitioned as
separate tasks. Consequently, a disease--phenotype association evaluated in
one direction could occur in training in the inverse direction.
Reciprocal-task performance therefore evaluates whether shared relational
supervision supports both retrieval formulations, rather than discovery of
previously unseen disease--phenotype associations.

Tier~3 held out exact phenotype-triplet--disease relevance pairs and
same-task positive disease targets. Most evaluation triplet queries were also
absent from training; however, 146 of 2,537 queries (5.8\%) occurred in
training with different disease targets, and constituent phenotype--disease
associations could appear in the relational training tasks. OntologyBench
therefore evaluates concept grounding, transductive relational transfer, and
compositional retrieval of held-out phenotype-set--disease compatibilities
within a shared ontology vocabulary. It does not evaluate inductive
generalization to entirely unseen entities or relation types.

\subsection{Dataset Statistics}
\label{sec:dataset_statistics}

Table~\ref{tab:ontologybench_stats} summarizes the benchmark statistics across all retrieval tasks. AvgPos denotes the average number of ground-truth relevant (positive) target documents associated with each query. Additional dataset statistics are provided in Appendix~\ref{para:t1a}.

Tier~1 tasks are one-to-one concept grounding (AvgPos = 1), while Tier~2 ontology association tasks exhibit substantial ambiguity due to many-to-many biomedical relationships (e.g., Dis→Phen and Phen→Dis average 19.8 and 16.6 positives per query). In contrast, Tier~3 compositional queries correspond to a smaller relevant set (4.39 diseases on average), reflecting the narrowing effect of combining multiple phenotype constraints. Sequence length statistics are reported in Appendix~\ref{sec:appendix_token_length}.

\section{Method}

\subsection{Supervision Settings}
\label{sec:supervision}

To study how ontology supervision shapes embedding representations, we compare two training regimes using the same task-specific split artifacts.

\paragraph{Concept Grounding Supervision.}
Models are trained only on alias–definition grounding pairs that map surface biomedical expressions to canonical concepts. No ontology relations or compositional signals are provided during training. Performance on higher-tier tasks, therefore, measures whether relational structure can emerge from concept grounding alone.

\paragraph{Unified Ontology Supervision.}
Models are trained jointly on all ontology-derived tasks, including grounding, entity relations (e.g., phenotype–disease, disease–gene), and compositional phenotype queries. 

\subsection{Two-Stage Retrieval Evaluation}
\label{para:two_section}

Performance degradation under compositional retrieval may arise from either missing candidates during retrieval or incorrect downstream ranking among retrieved candidates. To distinguish candidate-availability failure from downstream-ranking performance, we evaluated the two stages separately. Candidate availability was measured over the complete query set.

First, an initial embedding retriever generates the top-50 candidates from the full corpus. At least one relevant disease is present among these candidates for 82.1\% of queries (Hit@50 = 0.821), establishing the maximum proportion of queries for which downstream reranking can succeed. Here, Hit@k denotes the proportion of queries for which at least one relevant candidate appears among the top-k results. Second, we evaluate whether rerankers can place at least one relevant disease near the top of this fixed candidate set. Late-interaction reranking uses GTE-ModernColBERT with ColBERT-style MaxSim scoring, whereas LLM-based candidate scoring directly evaluates phenotype--disease compatibility for each retrieved candidate (Appendix~\ref{appendix:prompt}).

Candidate-generation metrics identify queries for which reranking could not recover a relevant disease because none was available. End-to-end metrics retain all queries and therefore characterize the combined two-stage system.

\subsection{Experimental Setup}
\paragraph{Model Setup.}

We evaluate lexical, dense retrieval, reranking, and generative models. As a lexical baseline, we use BM25. Dense retrieval models include general-purpose embedding models such as all-MiniLM-L6-v2~\citep{reimers2019sentence}, Qwen3 embeddings (0.6B and 4B)~\citep{zhang2025qwen3}, and OpenAI \texttt{text-embedding-3-large}\footnote{\url{https://platform.openai.com/docs/guides/embeddings}}, as well as biomedical embedding models including BioLORD-0.1B~\citep{remy-etal-2023-biolord}, MedEmbed-0.1B~\citep{balachandran2024medembed}, and SapBERT-0.1B~\citep{liu-etal-2021-self}.

For late-interaction reranking, we evaluate GTE-ModernColBERT~\citep{chaffin2025pylate} using ColBERT-style MaxSim scoring~\citep{khattab2020colbert}. For LLM-based candidate scoring reranking, we use Qwen3-4B-Thinking-2507, Qwen3-4B-Instruct-2507~\citep{qwen3technicalreport}, Qwen3.6-27B~\citep{qwen3.6-27b}, Gemma4-31B~\citep{google_gemma4_31b_2026}, and OpenAI GPT5.4-mini\footnote{\url{https://developers.openai.com/api/docs/guides/reasoning}}. 

\paragraph{Ontology-Aware Reference Methods.}

We include ontology-aware reference methods to estimate retrieval performance when curated biomedical structure is explicitly available. These methods operate under a different information regime from text-only retrievers and are therefore interpreted as diagnostic references rather than directly comparable baselines.

For Dis→Phen, Phen→Dis, and PhenTriplet→Dis, we use Phenomizer-style information-content similarity with set-level aggregation over HPO phenotypes. \citep{kohler2009clinical} For Phen→Gene, we use graph traversal over phenotype–disease–gene paths in the train-only inductive graph. This graph contains 350 of the 447 unique Phen→Gene evaluation targets, and 1,484 of 1,507 queries can reach at least one candidate gene through an indirect path.

Note that we do not report an ontology-aware reference for Tier 1 concept grounding. These tasks require mapping textual aliases to canonical concept definitions, whereas ontology traversal requires an already grounded concept identifier. Resolving the alias through the ontology’s alias table would directly disclose the target concept and reduce the task to a tautological dictionary lookup. Also, Dis→Gene result was not reported because the same inductive graph contains none of the unique evaluation target genes. The absence of target-gene coverage prevents the method from assigning meaningful graph-derived scores. We exclude the transductive graph because it contains the held-out disease–gene edges and reduces retrieval to direct lookup.

\paragraph{Training Objective.}
Fine-tuned models use architecture-specific contrastive objectives.
BioLORD-0.1B uses Cached Multiple Negatives Ranking Loss
(CachedMNRL), while Qwen3-Embedding-0.6B uses InfoNCE.
ModernColBERT uses the PyLate CachedContrastive objective with
ColBERT-style MaxSim scoring. These objectives share the general
softmax contrastive form
\begin{equation}
\mathcal{L}(q,d^+)
=
-\log
\frac{\exp(\operatorname{sim}(q,d^+))}
{\sum_{d\in\mathcal{B}}
 \exp(\operatorname{sim}(q,d))},
\end{equation}
where $\mathcal{B}$ denotes the contrastive candidate set. Similarity
is computed using cosine similarity for the bi-encoder models and
MaxSim aggregation for ModernColBERT. Architecture-specific loss and
batching configurations are reported in Appendix~\ref{appendix:optimization}.

\paragraph{Training Setup.}
Within each architecture, the Tier~1 and unified-supervision
configurations use the same epoch count and architecture-specific
optimization settings. All fine-tuned configurations were trained for
two epochs on one NVIDIA H100 MIG device. Complete batching,
optimization, and run-accounting details are provided in
Appendix~\ref{appendix:optimization}.

\subsection{Evaluation Metrics}

Retrieval performance is evaluated using normalized discounted cumulative gain (nDCG), mean reciprocal rank (MRR), and Hit@k, where \(k \in \{1,5,10\}\). We report nDCG@10 in the main paper and provide additional metrics and implementation details in Appendix~\ref{evaluation_protocol}. Relevance is binary, and metrics are computed over the complete candidate set for each task and macro-averaged across queries. Although we use standard retrieval metrics, ontology structure is incorporated through task construction: queries, candidate sets, and relevance labels are derived from curated biomedical relations and compositional phenotype structures. The metrics therefore assess retrieval under ontology-constrained relevance rather than ontology consistency directly.

\begin{table*}[t]
  \centering
  \small
  \resizebox{\textwidth}{!}{
  \begin{tabular}{lccc|cccc|c}
  \toprule
  & \multicolumn{8}{c}{nDCG@10} \\
  \cmidrule(lr){2-9}
  & \multicolumn{3}{c}{Tier~1}
  & \multicolumn{4}{c}{Tier~2}
  & Tier~3 \\
  \cmidrule(lr){2-4}
  \cmidrule(lr){5-8}
  Model
  & Gene$\rightarrow$Def
  & Phen$\rightarrow$Def
  & Dis$\rightarrow$Def
  & Dis$\rightarrow$Phen
  & Phen$\rightarrow$Dis
  & Phen$\rightarrow$Gene
  & Dis$\rightarrow$Gene
  & PhenTriplet$\rightarrow$Dis \\
  \midrule

  BM25
  & 0.087 & 0.381 & 0.275
  & 0.070 & 0.103 & 0.071 & 0.054
  & 0.107 \\

  MiniLM-L6-v2
  & 0.263 & 0.629 & 0.314
  & 0.106 & 0.153 & 0.109 & 0.058
  & 0.123 \\

  MedEmbed-0.1B
  & 0.440 & 0.738 & 0.454
  & 0.130 & 0.169 & 0.134 & 0.068
  & 0.143 \\

  SapBERT-0.1B
  & 0.670 & 0.818 & 0.476
  & 0.088 & 0.159 & 0.113 & 0.069
  & 0.131 \\

  BioLORD-0.1B
  & 0.290 & 0.807 & 0.426
  & 0.091 & 0.178 & 0.105 & 0.058
  & 0.151 \\

  \ftmodel{BioLORD-0.1B†}
  & 0.513 & \textbf{0.869} & \textbf{0.562}
  & 0.112 & 0.175 & 0.091 & 0.048
  & 0.171 \\

  \ftmodel{BioLORD-0.1B‡}
  & 0.455 & 0.668 & 0.532
  & 0.160 & 0.166 & 0.191 & 0.089
  & 0.244 \\

  \midrule

  Qwen3-Embed-0.6B
  & 0.325 & 0.605 & 0.247
  & 0.089 & 0.151 & 0.106 & 0.063
  & 0.109 \\

  \ftmodel{Qwen3-Embed-0.6B†}
  & 0.615 & 0.864 & 0.538
  & 0.112 & 0.158 & 0.076 & 0.045
  & 0.158 \\

  \ftmodel{Qwen3-Embed-0.6B‡}
  & 0.499 & 0.572 & 0.488
  & \textbf{0.184} & \textbf{0.283} & \textbf{0.269}
  & \textbf{0.110} & \textbf{0.313} \\

  Qwen3-Embed-4B
  & 0.526 & 0.616 & 0.330
  & 0.119 & 0.188 & 0.129 & 0.073
  & 0.080 \\

  \midrule

  OpenAI (text-embedding-3-large)
  & \textbf{0.747} & 0.860 & 0.516
  & 0.116 & 0.195 & 0.136 & 0.069
  & 0.169 \\

  \bottomrule
  \end{tabular}
  }
  \caption{Retrieval performance on OntologyBench, evaluated using nDCG@10
  across eight retrieval tasks spanning three retrieval tiers from grounding
  to compositional retrieval. Fine-tuned models are highlighted in blue, where
  \ftmodel{†} indicates models fine-tuned on the Tier~1 training set only, and
  \ftmodel{‡} indicates models fine-tuned jointly on the training sets from all
  three tiers.}
  \label{tab:ontologybench_results}
  \end{table*}

\section{Results}

\subsection{Ontology-Aware Retrieval Reveals a Gap in Compositional Compatibility}

Ontology-aware reference methods demonstrate that OntologyBench retrieval tasks are highly solvable when explicit biomedical structure is directly accessible. As shown in Table~\ref{tab:ontology-ndcg_baselines}, ontology-aware methods operating over curated hierarchical and relational structure achieve strong retrieval performance across phenotype-based and compositional tasks (e.g., PhenTriplet$\rightarrow$Dis nDCG@10 = 0.739). These ontology-aware reference methods operate under fundamentally different assumptions from embedding-based retrieval models. Rather than inferring compatibility from textual representations, they directly exploit explicit ontology hierarchy or curated disease--gene connectivity. The ontology-aware reference methods are therefore not intended as direct competitors, but instead serve as diagnostic references establishing the level of retrieval performance achievable when structured biomedical compatibility is explicitly available.

The ontology-aware reference methods reveal a substantial performance gap relative to the evaluated embedding retrieval systems. While ontology-aware reference methods directly exploit curated biomedical relationships and hierarchical structure, embedding models must recover such compatibility implicitly through representation learning from unstructured text and retrieval supervision alone. The observed performance gap suggests that current embedding training objectives may not fully preserve ontology-defined compatibility.

\begin{table}[t]
\centering
\small
\setlength{\tabcolsep}{5pt}
\begin{tabular}{llc}
\toprule
Task & Method & nDCG@10 \\
\midrule
Dis$\rightarrow$Phen
& Phenomizer-style & 0.482 \\

Phen$\rightarrow$Dis
& Phenomizer-style & 0.788 \\

PhenTriplet$\rightarrow$Dis
& Phenomizer-style & 0.739 \\

Phen$\rightarrow$Gene
& Graph traversal & 0.694 \\
\bottomrule
\end{tabular}
\caption{Performance of ontology-aware reference methods. Phenotype--disease and compositional tasks use Phenomizer-style ontology-based semantic similarity, while Phen$\rightarrow$Gene uses graph traversal over phenotype--disease--gene paths.}
\label{tab:ontology-ndcg_baselines}
\end{table}

\subsection{Retrieval Performance Differs Across Grounding, Relational, and Compositional Tasks}

Table~\ref{tab:ontologybench_results} shows substantial variation across retrieval tasks. Several models achieve their highest scores on concept-grounding tasks, whereas scores are generally lower on relational and compositional tasks. These cross-tier differences are descriptive because candidate spaces, relevance structures, and query ambiguity also vary across tasks. On the fixed PhenTriplet→Dis task, however, the best evaluated embedding model achieves nDCG@10 of 0.313, compared with 0.739 for the ontology-aware baseline.

\subsection{Task-Dependent Effects of Domain Specification and Supervision}

Domain specialization provides an advantage in some comparisons but not uniformly across relational tasks. On Phenotype-Triplet$\rightarrow$Disease, BioLORD-0.1B achieves nDCG@10 = 0.151, exceeding the pretrained Qwen3-Embed-0.6B and Qwen3-Embed-4B scores of 0.109 and 0.080, respectively.

Unified multi-tier supervision produces substantial gains on several relational and compositional tasks. Relative to its pretrained counterpart, Qwen3-Embed-0.6B improves by +0.047 to +0.163 nDCG@10 across the four Tier~2 tasks and by +0.204 on Phenotype-Triplet$\rightarrow$Disease. It also exceeds the strongest pretrained baseline by up to +0.133 across Tier~2 and by +0.144 on Phenotype-Triplet$\rightarrow$Disease. Some model–task combinations decline after fine-tuning, however, indicating that the benefit is not universal. Within the evaluated configurations, task-aligned supervision can produce larger gains than increasing model size, but these results do not establish a general ordering between supervision and scale. Additional evaluation results are provided in Appendix~\ref{appendix:additional}.

\subsection{Late Interaction and Generative Reranking Provide Limited End-to-End Improvement}

Late-interaction rerankers substantially underperform the first-stage retriever on Tier-3 compositional retrieval tasks. Tier~1 only fine-tuning produced little improvement in ModernColBERT performance, whereas unified multi-tier fine-tuning reduced its performance (Table~\ref{tab:tier3_reasoning}). LLM-based candidate scoring performed better than the evaluated token-interaction rerankers but the strongest LLM scorers produced only small and inconsistent differences relative to the first-stage retriever. Qwen3.6-27B increased MRR@10 from 0.297 to 0.306 and nDCG@10 from 0.313 to 0.315, while its Hit@10 was slightly lower. The strongest LLM scorers produced performance comparable to, rather than clearly exceeding, the retriever. GPT-5.4-mini produced the highest Hit@10, but its MRR@10 and
nDCG@10 remained comparable to the retriever.

The first-stage retriever returned at least one relevant disease in its top-50 candidates for 2,083 of 2,537 queries (82.1\%). For the remaining 454 queries (17.9\%), reranking could not recover a relevant disease because no relevant candidate was available to reorder. These queries were retained in the evaluation and received zero retrieval credit, so the reported metrics reflect the performance of the complete two-stage system. Because paired uncertainty estimates were not calculated, the small numerical differences among rerankers should not be interpreted as evidence that one method was superior. Overall, none of the evaluated rerankers clearly improved on the first-stage retriever, indicating that reranking did not resolve the Tier-3 retrieval bottleneck.

\begin{table}[t]
    \centering
    \small
    \setlength{\tabcolsep}{5pt}
    \renewcommand{\arraystretch}{0.6}

    \begin{tabular}{lccc}
    \toprule
    & \multicolumn{3}{c}{$k = 10$} \\
    \cmidrule(lr){2-4}
    \textbf{Model} & \textbf{MRR} & \textbf{Hit} & \textbf{nDCG} \\
    \midrule

    \textit{Retriever} \\
    \ftmodel{Qwen3-Embed-0.6B$^{\ddagger}$}
    & 0.297 & 0.572 & 0.313 \\

    \midrule
    \textit{Token-Interaction Rerankers} \\
    ModernColBERT
    & 0.168 & 0.384 & 0.180 \\
    \ftmodel{ModernColBERT$^{\dagger}$}
    & 0.168 & 0.386 & 0.183 \\
    \ftmodel{ModernColBERT$^{\ddagger}$}
    & 0.135 & 0.384 & 0.163 \\

    \midrule
    \textit{LLM Candidate Scoring} \\
    Qwen3-4B-Instruct-2507
    & 0.262 & 0.525 & 0.275 \\
    Qwen3-4B-Thinking-2507
    & 0.278 & 0.546 & 0.288 \\
    Qwen3.6-27B
    & \textbf{0.306} & 0.571 & \textbf{0.315} \\
    Gemma4-31B
    & 0.284 & 0.524 & 0.287 \\
    OpenAI (GPT5.4-mini)
    & 0.301 & \textbf{0.574} & 0.312 \\

    \bottomrule
    \end{tabular}
 \caption{End-to-end performance on the Tier-3
    PhenTriplet$\rightarrow$Dis task ($N=2{,}537$). MRR@10, binary
    Hit@10, and nDCG@10 were calculated over the complete query set.
    The first-stage top-50 candidate sets contained at least one
    relevant disease for 2,083 queries (82.1\%); the remaining 454
    queries remained in the denominator and contributed zero retrieval
    credit. Candidate-scoring methods reordered retrieved candidates
    but did not introduce additional diseases.}

    \label{tab:tier3_reasoning}
\end{table}

\subsection{Failure Modes in Compositional Retrieval}

We perform a rule-based error analysis (Appendix~\ref{sup:error_analysis}) of Tier-3 queries
for which the embedding model does not rank the ground-truth disease
first. The analysis distinguishes failures for which the ontology-aware
reference retrieves the ground-truth disease within the top 10 from
remaining failures. Failures cluster into four modes: (1) \textbf{aggregation failure} (58.5\%), where the model retrieves diseases matching only a subset of the query phenotypes rather than the full phenotype combination; (2) \textbf{generic phenotype bias} (25.0\%), where retrieval is driven by broad or non-specific phenotypes; (3) \textbf{related-disease confusion} (11.8\%), where the predicted disease is phenotypically similar to the ground-truth disease; and (4) \textbf{semantic drift} (4.7\%), where retrieved diseases show low phenotype overlap and low ontology similarity. Aggregation failures dominate, indicating that embedding models often match phenotypes independently rather than enforcing joint compatibility across the full phenotype set. Unified multi-tier supervision improves Tier~3 retrieval substantially, increasing Hit@10 from 0.170 to 0.572 and nDCG@10 from 0.109 to 0.313 (Appendix~\ref{appendix:additional} Table~\ref{tab:tier3-combined}). However, performance remains below the ontology-aware reference, indicating that compositional ranking errors remain common.

The evaluated embedding models frequently retrieved diseases compatible with only part of the phenotype set. This finding characterizes the evaluated systems and task construction; they do not establish that all embedding architectures are inherently limited to represent compositional biomedical information. This limitation is consistent with lower performance on relation types (e.g., phenotype→gene), indicating that compositional structure remains underrepresented in current representations.

\section{Discussion}

Retrieval systems are increasingly expected to recover compatibility across multiple structured constraints rather than simply retrieve semantically similar text. OntologyBench evaluates this distinction across concept-grounding, relational, and compositional retrieval tasks. Across the evaluated architectures, performance is generally lower on relational and compositional tasks than on several grounding tasks. These cross-task differences are descriptive because candidate spaces, relevance structures, and query ambiguity also vary. Within the fixed Tier-3 task, however, the evaluated embedding models remain substantially below the ontology-aware reference method, indicating that the evaluated text-based configurations do not recover ontology-defined relevance as effectively when multiple phenotype constraints must be combined.

Unified multi-tier supervision improves performance on several relational and compositional tasks, indicating that ontology-derived supervision can produce representations that better support these retrieval relationships. However, the remaining gap relative to ontology-aware reference methods cannot be attributed solely to the embedding objectives. The reference methods directly access curated ontology relations, whereas the embedding models operate over textual representations that may not contain all information used to define relevance. The results therefore motivate retrieval architectures that integrate textual representations with ontology knowledge rather than establishing that embedding objectives alone are responsible for the observed gap.

Our two-stage evaluation provides additional insight into where performance is lost. Under the evaluated configurations, token-interaction reranking underperforms the first-stage retriever, while the strongest LLM-based candidate scorer provides only a modest end-to-end improvement over the fixed retriever. Because these metrics include queries for which no relevant disease appears in the initial candidate set, they reflect both candidate-generation and downstream-ranking limitations. Within the rule-selected Tier-3 failure subset, the largest category consists of predictions matching only part of the phenotype combination. Together, these findings suggest that improving downstream scoring alone may be insufficient when relevant candidates are omitted or compatibility is not adequately represented during initial retrieval.

Although OntologyBench is constructed from biomedical ontologies, its evaluation framework may be applicable beyond biomedicine. Retrieval-augmented generation, agentic systems, knowledge-graph retrieval, and other knowledge-intensive NLP applications may require candidates satisfying multiple structured constraints rather than maximizing semantic similarity alone. By organizing tasks across grounding, relational, and compositional retrieval, OntologyBench provides a framework for examining retrieval behaviour under different forms of structured compatibility. We hope it supports future work on retrieval objectives and architectures that more effectively integrate representation learning with structured knowledge.

\section{Conclusion}

We introduced OntologyBench, a benchmark spanning concept grounding, ontology-relation retrieval, and phenotype-set retrieval. Scores were generally lower on relational and compositional tasks than on several grounding tasks. On the fixed PhenTriplet→Dis task, the evaluated embedding configurations remained below the ontology-aware reference methods, and the evaluated rerankers provided little or no end-to-end improvement over the first-stage retriever. Candidate omission contributed to the observed end-to-end gap. 

These findings motivate retrieval systems that combine flexible textual representation with explicit structured knowledge; they do not establish a universal ordering between embedding and symbolic approaches.

\section{Data and Code Availability}

The OntologyBench benchmark is publicly available on
Github\footnote{\url{https://github.com/cindyzhangxy/OntologyBench}} under the MIT license,
with the benchmark dataset additionally distributed through
Hugging Face\footnote{\url{https://huggingface.co/datasets/cxyzhang/OntologyBench}}
under the CC BY 4.0 License.

\section*{Limitations}

OntologyBench evaluates retrieval under controlled biomedical compatibility constraints and does not capture broader aspects of real-world diagnosis, such as temporal progression, causal mechanisms, incomplete observations, or patient-specific context.

Task-specific partitioning prevents overlap of exact query–document pairs and same-task target identifiers. However, biomedical concepts and underlying relations may recur across training tasks. The unified setting therefore evaluates transductive multi-task transfer within a shared ontology vocabulary rather than generalization to entirely unseen entities or relations. However, scores were generally lower on relational and compositional tasks than on several grounding tasks, although task characteristics differ and these cross-tier contrasts should not be interpreted as isolated effects of constraint complexity (Table~\ref{tab:ontologybench_results}). 

Our experiments evaluate a representative range of retrieval architectures and model scales, but are not intended as an exhaustive leaderboard-style comparison of all emerging foundation systems. The primary goal of OntologyBench is instead to characterize retrieval behaviour across grounding, relational, and compositional retrieval tasks. Future work may investigate whether larger-scale or ontology-aware architectures mitigate the limitations identified here.

OntologyBench evaluates retrieval behaviour under structured biomedical constraints rather than directly measuring preservation of ontology geometry within embedding space. Standard retrieval metrics such as nDCG, MRR, and Hit@k therefore do not explicitly quantify hierarchical distance preservation or geometric alignment with ontology topology. Consequently, the lower scores observed on relational and compositional retrievals should be interpreted as evidence of limitations in compositional retrieval rather than a definitive measure of ontology structure preservation within learned representations.

Although ontology-aligned supervision improves relational and compositional retrieval, the current study does not fully isolate which components of the supervision signal contribute most strongly to these gains. Improvements may arise from a combination of structured relational supervision, task-specific retrieval supervision, and domain adaptation effects. Future work could further disentangle these factors through more controlled supervision ablations.

Finally, OntologyBench focuses on representation-level retrieval. While practical clinical systems may incorporate reranking, generative reasoning, and expert validation, our results suggest that substantial limitations already emerge at the retrieval stage. The benchmark should therefore be interpreted as a diagnostic framework for analyzing retrieval behaviour under structured biomedical constraints rather than a direct proxy for end-to-end clinical performance.

\section*{Ethical Considerations}
OntologyBench is constructed entirely from publicly available biomedical ontologies released under open licenses, including the HPO and MONDO (CC-BY 4.0) and HGNC and NCBI resources (public domain). These sources provide structured definitions of concepts and relations rather than patient-level data. OntologyBench contains no personally identifiable information or sensitive health records.

\section*{Statement of AI Use}
Generative AI was used for language editing. All research design, methodology, experiments, analyses, results, and conclusions are the work of the authors and were verified by the authors.

\section*{Acknowledgments}
This research was enabled in part through the computational resources provided by Advanced Research Computing at the University of British Columbia and the Digital Research Alliance of Canada. This work was funded by the Nenad Blau IEMBase Endowment Fund of the MCF, Marin County, CA, USA. This work was also supported by a Natural Sciences and Engineering Research Council of Canada (NSERC) Discovery Grant (RGPIN-2024-06783) awarded to WWW; a BC Children's Hospital Research Institute (BCCHR) Doctoral Studentship awarded to XYCZ; and an NSERC Discovery Grant and a Canada Foundation for Innovation John R. Evans Leaders Fund (CFI-JELF) grant awarded to JZ.

\bibliography{anthology,custom}
\bibliographystyle{acl_natbib}

\appendix

\section{Dataset Examples}
\label{appendix:examples}

This appendix shows representative examples from the three tiers of
OntologyBench tasks.

\subsection{Tier~1: Concept Grounding}

\begin{tcolorbox}[title=Gene→Document Retrieval]
{\ttfamily
doc\_id: GENE:2011

doc: Protein-coding gene located on chromosome 11q13.1 belonging to the
microtubule affinity regulating kinase family. The encoded kinase
regulates epithelial and neuronal cell polarity and microtubule stability.
}
\end{tcolorbox}

\begin{tcolorbox}[title=Phen→Definition Retrieval]
{\ttfamily
doc\_id: HP:0006349

doc: Congenital absence of one or more permanent teeth, including
hypodontia, oligodontia, or complete tooth absence.
}
\end{tcolorbox}

\begin{tcolorbox}[title=MONDO→Definition Retrieval]
{\ttfamily
doc\_id: MONDO:0014543

doc: Congenital myasthenic syndrome with glycosylation defect
caused by mutations in the ALG2 gene.
}
\end{tcolorbox}

\subsection{Tier~2: Ontology-Level Relations}

\begin{tcolorbox}[title=Disease→Phenotype]
{\ttfamily
doc\_id: HP:0002154

doc: Elevated concentration of glycine in the blood.
}
\end{tcolorbox}

\begin{tcolorbox}[title=Phenotype→Disease]
{\ttfamily
doc\_id: MONDO:0009479

doc: Johanson-Blizzard syndrome characterized by exocrine pancreatic
insufficiency, nasal alae hypoplasia, hearing loss, growth retardation,
and intellectual disability.
}
\end{tcolorbox}

\begin{tcolorbox}[title=Phenotype→Gene]
{\ttfamily
doc\_id: GENE:80185

doc: TTI2 encodes a regulator of the DNA damage response and a
component of the Triple T complex involved in cellular resistance
to DNA damage stresses.
}
\end{tcolorbox}

\begin{tcolorbox}[title=Disease→Gene]
{\ttfamily
doc\_id: GENE:889

doc: KRIT1 encodes a protein involved in beta1-integrin-mediated
cell proliferation and endothelial junction integrity. Mutations
cause cerebral cavernous malformations.
}
\end{tcolorbox}

\subsection{Tier~3: Compositional Phenotype-Based Disease Retrieval}

\begin{tcolorbox}[title=Multi-Phenotype→Disease]
{\ttfamily
doc\_id: MONDO:0016394

doc: Sporadic infantile bilateral striatal necrosis characterized by
degeneration of the caudate nucleus, putamen, and globus pallidus,
leading to developmental regression and movement disorders.
}
\end{tcolorbox}

\begin{tcolorbox}[title=Multi-Phenotype→Disease]
{\ttfamily
doc\_id: MONDO:0011614

doc: HMG-CoA synthase-2 deficiency, a rare disorder of ketone body
metabolism presenting with vomiting, lethargy, hepatomegaly,
and non-ketotic hypoglycemia.
}
\end{tcolorbox}

\begin{tcolorbox}[title=Multi-Phenotype→Disease]
{\ttfamily
doc\_id:  MONDO:0020502

doc: Yellow fever, a zoonotic viral disease that may progress from
fever and systemic symptoms to hemorrhagic fever and multi-organ failure.
}
\end{tcolorbox}

\section{Additional Dataset Analysis}
\subsubsection*{Tier~1 Statistics}
\label{para:t1a}

Disease identity grounding (MONDO) contains 6,611 canonical concepts associated with 48,854 alias queries (mean 7.39 aliases per concept; median 6; maximum 50). 

Alias polysemy is limited: 1.0\% of surface forms map to more than one concept. Gene identity grounding contains 2,345 gene concepts, each associated with a single canonical symbol. Phenotype grounding includes 8,253 phenotype concepts with extensive synonym coverage.

These statistics indicate that Tier~1 primarily evaluates lexical normalization under moderate alias diversity rather than heavy ambiguity.

\subsubsection*{Phenotype Cardinality and Hypothesis Space Reduction}
\label{tier3:cardinality}
We fix Tier-3 phenotype cardinality to three to study the minimal non-trivial regime of compositional retrieval. Ontology association statistics indicate:

\begin{itemize}
    \item Single phenotype: mean 328.9 compatible diseases
    \item Phenotype pairs: mean 26.7 compatible diseases
    \item Phenotype triplets: mean 5.0 compatible diseases
    \item Phenotype quadruplets: mean 1.95
    \item Phenotype quintuplets: mean 1.31
\end{itemize}

Triplets provide substantial hypothesis-space reduction while avoiding near-deterministic collapse observed at higher cardinalities.

\section{Experimental Details}

\subsubsection*{Sequence Length Distribution and Truncation}
\label{sec:appendix_token_length}

\begin{table}[!htbp]
\centering
\small
\begin{tabular}{lrrrr}
\toprule
\textbf{Model} & \textbf{Mean} & \textbf{P95} & \textbf{P99} & \textbf{Max} \\
\midrule
Qwen3-Embed-0.6B & 33.40 & 112 & 161 & 1053 \\
BioLORD-0.1B & 48.39 & 128 & 205 & 926 \\
ColBERT & 25.08 & 88 & 149 & 845 \\
\bottomrule
\end{tabular}
\caption{Token length statistics of training instances under different tokenizers (no truncation applied). We report the mean length, 95th percentile (P95), 99th percentile (P99), and maximum sequence length.}
\end{table}

\label{sec:token_length}

Bi-encoder models used a maximum sequence length of 256 tokens to
limit truncation under the heavy-tailed definition-length distribution.
ModernColBERT retained its architecture-specific maximum lengths of
48 query tokens and 300 document tokens. Optimization, batching,
hardware, and run-accounting details are provided in
Appendix~\ref{appendix:optimization}.

\section{Optimization and Training Setup}
\label{appendix:optimization}

All fine-tuned configurations were trained for two epochs on one NVIDIA
H100 MIG device with gradient accumulation of 1 and random seed 42.
Reported fine-tuning results are from single training runs rather than
averages across multiple seeds.

BioLORD-0.1B was fine-tuned with Cached Multiple Negatives Ranking Loss
using a batch size of 256 and cached forward mini-batches of 128.
Qwen3-Embedding-0.6B was fine-tuned with InfoNCE using a batch size of
256 packed records, sequence packing with a length of 256 tokens, and
no cached-loss subdivision. ModernColBERT was fine-tuned with the
PyLate CachedContrastive objective using a batch size of 256 and cached
forward mini-batches of 32. Cross-device negative gathering was
disabled for cached-loss configurations.

Bi-encoder models used a maximum sequence length of 256 tokens.
ModernColBERT used maximum query and document lengths of 48 and 300
tokens, respectively. Learning rates were $2\times10^{-5}$ for
BioLORD-0.1B, $5\times10^{-5}$ for Qwen3-Embedding-0.6B, and
$3\times10^{-6}$ for ModernColBERT. Complete training configurations
and saved trainer states are included in the released repository.

\section{Evaluation Protocol}
\label{evaluation_protocol}

For each task, every query is evaluated by ranking all documents in the corresponding task-specific corpus without heuristic filtering or ontology pruning. Metrics are computed independently for each task at $k \in \{1,5,10\}$ and macro-averaged across all test queries.

Let $G_q$ denote the non-empty set of relevant documents for query $q$, and let $d_{q,i}$ denote the document at rank $i$ after duplicate document identifiers have been removed. Relevance is binary:

\begin{equation}
\operatorname{rel}_{q,i}
=
\begin{cases}
1, & d_{q,i} \in G_q,\\
0, & d_{q,i} \notin G_q.
\end{cases}
\end{equation}

Discounted cumulative gain at cutoff $k$ is defined as

\begin{equation}
\operatorname{DCG}_q@k
=
\sum_{i=1}^{k}
\frac{\operatorname{rel}_{q,i}}
{\log_2(i+1)}.
\end{equation}

For binary relevance, the ideal ranking contains
$\min(k,|G_q|)$ relevant documents. Therefore,

\begin{equation}
\operatorname{IDCG}_q@k
=
\sum_{i=1}^{\min(k,|G_q|)}
\frac{1}{\log_2(i+1)},
\end{equation}

and normalized discounted cumulative gain is

\begin{equation}
\operatorname{nDCG}_q@k
=
\frac{\operatorname{DCG}_q@k}
{\operatorname{IDCG}_q@k}.
\end{equation}

Let $r_q$ denote the rank of the first relevant document. The truncated reciprocal-rank and Hit@$k$ metrics are

\begin{equation}
\operatorname{RR}_q@k
=
\begin{cases}
1/r_q, & r_q \leq k,\\
0, & r_q > k,
\end{cases}
\end{equation}

\begin{equation}
\operatorname{Hit}_q@k
=
\begin{cases}
1, & r_q \leq k,\\
0, & r_q > k.
\end{cases}
\end{equation}

For any query-level metric $m_q@k$, the reported task-level score is the macro-average over the complete test-query set $Q$:

\begin{equation}
M@k
=
\frac{1}{|Q|}
\sum_{q \in Q} m_q@k.
\end{equation}

Thus, MRR@$k$ is the macro-average of
$\operatorname{RR}_q@k$, while Hit@$k$ is the proportion of test queries with at least one relevant document among the top-$k$ retrieved results.
\

\onecolumn
\section{Prompt Used for Generative Candidate Scoring}
\label{appendix:prompt}

For LLM-based candidate scoring, each candidate disease was evaluated independently using the following prompt template. The same semantic prompt was supplied through each model's native chat template.  

\begin{tcolorbox}[
    colback=gray!5,
    colframe=black!40,
    boxrule=0.5pt,
    breakable,
    boxsep=2pt,
    left=3pt,
    right=3pt,
    top=3pt,
    bottom=3pt,
    before skip=4pt,
    after skip=4pt
]
\scriptsize
\footnotesize

\textbf{System instruction}

You are evaluating the textual compatibility between a candidate disorder description and a set of patient phenotypes.

Use ONLY the information provided in the phenotype list and the candidate disorder description. Do NOT use external medical knowledge.

\vspace{3pt}

\textbf{Patient phenotypes}

\begin{itemize}
    \item Increased lower facial height
    \item Laryngeal stenosis
    \item Long philtrum
\end{itemize}

\textbf{Candidate disorder description}

Autosomal dominant prognathism. Malocclusion in which the mandible is anterior to the maxilla, as reflected by the relationship of the first permanent molars.

\vspace{3pt}

\textbf{Task}

Assign an integer compatibility score from 0 to 100 indicating how well the candidate disorder description supports the complete phenotype set.

\vspace{3pt}

\textbf{Scoring guideline}

\begin{itemize}
    \item 0--19: incompatible
    \item 20--39: weak overlap
    \item 40--59: partial match
    \item 60--79: strong match
    \item 80--100: near-complete match
\end{itemize}

\textbf{Output rules}

Output exactly one JSON object containing only the score. Do not include explanations, Markdown, additional fields, or repeated output.

\begin{verbatim}
{"score": 0}
\end{verbatim}

\end{tcolorbox}

\section{Additional Evaluation Metrics}
\label{appendix:additional}
\clearpage

\begin{table*}[!t]
\centering
\scriptsize
\begin{tabular}{l|ccc|ccc|ccc}
\toprule
& \multicolumn{9}{c}{nDCG@k} \\
\cmidrule(lr){2-10}
& \multicolumn{3}{c|}{Gene$\rightarrow$Def}
& \multicolumn{3}{c|}{Phen$\rightarrow$Def}
& \multicolumn{3}{c}{Dis$\rightarrow$Def} \\
Model
& @1 & @5 & @10
& @1 & @5 & @10
& @1 & @5 & @10 \\
\midrule

BM25
& 0.079 & 0.083 & 0.087
& 0.265 & 0.361 & 0.381
& 0.200 & 0.262 & 0.275 \\

MiniLM-L6-v2
& 0.158 & 0.244 & 0.263
& 0.500 & 0.609 & 0.629
& 0.224 & 0.296 & 0.314 \\

\ftmodel{BioLORD-0.1B‡}
& 0.311 & 0.428 & 0.455
& 0.452 & 0.639 & 0.668
& 0.389 & 0.505 & 0.532 \\

\ftmodel{BioLORD-0.1B†}
& 0.364 & 0.489 & 0.513
& \textbf{0.767} & \textbf{0.862} & \textbf{0.869}
& \textbf{0.425} & \textbf{0.540} & \textbf{0.562} \\

BioLORD-0.1B
& 0.167 & 0.268 & 0.290
& 0.693 & 0.795 & 0.807
& 0.309 & 0.402 & 0.426 \\

MedEmbed-0.1B
& 0.320 & 0.424 & 0.440
& 0.600 & 0.722 & 0.738
& 0.350 & 0.433 & 0.454 \\

SapBERT-0.1B
& 0.510 & 0.654 & 0.670
& 0.699 & 0.806 & 0.818
& 0.353 & 0.452 & 0.476 \\

\midrule

\ftmodel{Qwen3-Embed-0.6B‡}
& 0.348 & 0.476 & 0.499
& 0.390 & 0.541 & 0.572
& 0.335 & 0.460 & 0.488 \\

\ftmodel{Qwen3-Embed-0.6B†}
& 0.480 & 0.597 & 0.615
& 0.762 & 0.855 & 0.864
& 0.401 & 0.515 & 0.538 \\

Qwen3-Embed-0.6B
& 0.220 & 0.305 & 0.325
& 0.475 & 0.586 & 0.605
& 0.174 & 0.232 & 0.247 \\

Qwen3-Embed-4B
& 0.378 & 0.501 & 0.526
& 0.476 & 0.592 & 0.616
& 0.214 & 0.306 & 0.330 \\

\midrule

OpenAI (text-embedding-3-large)
& \textbf{0.619} & \textbf{0.730} & \textbf{0.747}
& 0.751 & 0.851 & 0.860
& 0.391 & 0.493 & 0.516 \\

\bottomrule
\end{tabular}

\caption{Tier~1 ontology grounding tasks evaluated using nDCG@k.}
\end{table*}


\begin{table*}[!t]
\centering
\scriptsize
\begin{tabular}{l|ccc|ccc|ccc}
\toprule
& \multicolumn{9}{c}{MRR@k} \\
\cmidrule(lr){2-10}

& \multicolumn{3}{c|}{Gene$\rightarrow$Def}
& \multicolumn{3}{c|}{Phen$\rightarrow$Def}
& \multicolumn{3}{c}{Dis$\rightarrow$Def} \\

Model
& @1 & @5 & @10
& @1 & @5 & @10
& @1 & @5 & @10 \\
\midrule

BM25
& 0.079 & 0.082 & 0.083
& 0.265 & 0.333 & 0.341
& 0.200 & 0.244 & 0.250 \\

MiniLM-L6-v2
& 0.158 & 0.219 & 0.227
& 0.500 & 0.579 & 0.587
& 0.224 & 0.275 & 0.282 \\

\ftmodel{BioLORD-0.1B‡}
& 0.311 & 0.392 & 0.404
& 0.452 & 0.586 & 0.598
& 0.389 & 0.472 & 0.483 \\

\ftmodel{BioLORD-0.1B†}
& 0.364 & 0.454 & 0.464
& \textbf{0.767} & \textbf{0.836} & \textbf{0.839}
& \textbf{0.425} & \textbf{0.507} & \textbf{0.516} \\

BioLORD-0.1B
& 0.167 & 0.237 & 0.246
& 0.693 & 0.766 & 0.771
& 0.309 & 0.375 & 0.385 \\

MedEmbed-0.1B
& 0.320 & 0.395 & 0.402
& 0.600 & 0.688 & 0.695
& 0.350 & 0.409 & 0.418 \\

SapBERT-0.1B
& 0.510 & 0.614 & 0.620
& 0.699 & 0.777 & 0.782
& 0.353 & 0.423 & 0.433 \\

\midrule

\ftmodel{Qwen3-Embed-0.6B‡}
& 0.348 & 0.438 & 0.447
& 0.390 & 0.497 & 0.509
& 0.335 & 0.423 & 0.435 \\

\ftmodel{Qwen3-Embed-0.6B†}
& 0.480 & 0.565 & 0.573
& 0.762 & 0.830 & 0.834
& 0.401 & 0.482 & 0.491 \\

Qwen3-Embed-0.6B
& 0.220 & 0.282 & 0.290
& 0.475 & 0.554 & 0.562
& 0.174 & 0.215 & 0.221 \\

Qwen3-Embed-4B
& 0.378 & 0.466 & 0.477
& 0.476 & 0.558 & 0.568
& 0.214 & 0.279 & 0.289 \\

\midrule

OpenAI (text-embedding-3-large)
& \textbf{0.619} & \textbf{0.699} & \textbf{0.706}
& 0.751 & 0.825 & 0.829
& 0.391 & 0.464 & 0.473 \\

\bottomrule
\end{tabular}

\caption{Tier~1 ontology grounding tasks evaluated using MRR@k.}
\label{tab:tier1-mrr}
\end{table*}

\begin{table*}[!t]
\centering
\scriptsize
\begin{tabular}{l|ccc|ccc|ccc}
\toprule
& \multicolumn{9}{c}{Hit@k} \\
\cmidrule(lr){2-10}

& \multicolumn{3}{c|}{Gene$\rightarrow$Def}
& \multicolumn{3}{c|}{Phen$\rightarrow$Def}
& \multicolumn{3}{c}{Dis$\rightarrow$Def} \\

Model
& @1 & @5 & @10
& @1 & @5 & @10
& @1 & @5 & @10 \\

\midrule

BM25
& 0.079 & 0.088 & 0.100
& 0.265 & 0.448 & 0.508
& 0.200 & 0.315 & 0.357 \\

MiniLM-L6-v2
& 0.158 & 0.318 & 0.378
& 0.500 & 0.700 & 0.761
& 0.224 & 0.360 & 0.417 \\

\ftmodel{BioLORD-0.1B‡}
& 0.311 & 0.534 & 0.619
& 0.452 & 0.797 & 0.886
& 0.389 & 0.605 & 0.686 \\

\ftmodel{BioLORD-0.1B†}
& 0.364 & 0.592 & 0.666
& \textbf{0.767} & \textbf{0.937} & \textbf{0.958}
& \textbf{0.425} & \textbf{0.639} & \textbf{0.706} \\

BioLORD-0.1B
& 0.167 & 0.362 & 0.429
& 0.693 & 0.879 & 0.916
& 0.309 & 0.485 & 0.559 \\

MedEmbed-0.1B
& 0.320 & 0.508 & 0.559
& 0.600 & 0.821 & 0.873
& 0.350 & 0.505 & 0.569 \\

SapBERT-0.1B
& \textbf{0.510} & \textbf{0.775} & \textbf{0.821}
& 0.699 & 0.892 & 0.929
& 0.353 & 0.538 & 0.612 \\

\midrule

\ftmodel{Qwen3-Embed-0.6B‡}
& 0.348 & 0.592 & 0.664
& 0.390 & 0.676 & 0.771
& 0.335 & 0.569 & 0.656 \\

\ftmodel{Qwen3-Embed-0.6B†}
& 0.480 & 0.689 & 0.745
& 0.762 & 0.929 & 0.955
& 0.401 & 0.615 & 0.684 \\

Qwen3-Embed-0.6B
& 0.220 & 0.374 & 0.436
& 0.475 & 0.681 & 0.741
& 0.174 & 0.283 & 0.329 \\

Qwen3-Embed-4B
& 0.378 & 0.606 & 0.680
& 0.476 & 0.692 & 0.768
& 0.214 & 0.386 & 0.461 \\

\midrule

OpenAI (text-embedding-3-large)
& \textbf{0.619} & \textbf{0.824} & \textbf{0.876}
& 0.751 & 0.928 & 0.956
& 0.391 & 0.582 & 0.652 \\

\bottomrule
\end{tabular}

\caption{Tier~1 ontology grounding tasks evaluated using Hit@k.}
\label{tab:tier1-hit}
\end{table*}


\begin{table*}[t]
\centering
\scriptsize
\begin{tabular}{l|ccc|ccc|ccc|ccc}
\toprule
& \multicolumn{12}{c}{nDCG@k} \\
\cmidrule(lr){2-13}
& \multicolumn{3}{c|}{Dis$\rightarrow$Gene}
& \multicolumn{3}{c|}{Dis$\rightarrow$Phen}
& \multicolumn{3}{c|}{Phen$\rightarrow$Dis}
& \multicolumn{3}{c}{Phen$\rightarrow$Gene} \\
Model
& @1 & @5 & @10
& @1 & @5 & @10
& @1 & @5 & @10
& @1 & @5 & @10 \\
\midrule

BM25
& 0.076 & 0.050 & 0.054
& 0.084 & 0.066 & 0.070
& 0.099 & 0.099 & 0.103
& 0.048 & 0.065 & 0.071 \\

MiniLM-L6-v2
& 0.079 & 0.055 & 0.058
& 0.120 & 0.098 & 0.106
& 0.150 & 0.143 & 0.153
& 0.058 & 0.095 & 0.109 \\

\ftmodel{BioLORD-0.1B‡}
& 0.112 & 0.083 & 0.089
& 0.158 & 0.142 & 0.160
& 0.034 & 0.119 & 0.166
& 0.104 & 0.165 & 0.191 \\

\ftmodel{BioLORD-0.1B†}
& 0.061 & 0.043 & 0.048
& 0.130 & 0.104 & 0.112
& 0.151 & 0.159 & 0.175
& 0.054 & 0.078 & 0.091 \\

BioLORD-0.1B
& 0.066 & 0.053 & 0.058
& 0.099 & 0.083 & 0.091
& 0.158 & 0.161 & 0.178
& 0.060 & 0.094 & 0.105 \\

MedEmbed-0.1B
& 0.095 & 0.066 & 0.068
& 0.148 & 0.120 & 0.130
& 0.155 & 0.157 & 0.169
& 0.079 & 0.118 & 0.134 \\

SapBERT-0.1B
& 0.096 & 0.066 & 0.069
& 0.104 & 0.080 & 0.088
& 0.142 & 0.148 & 0.159
& 0.068 & 0.101 & 0.113 \\

\midrule

\ftmodel{Qwen3-Embed-0.6B‡}
& \textbf{0.137} & \textbf{0.102} & \textbf{0.110}
& \textbf{0.185} & \textbf{0.165} & \textbf{0.184}
& \textbf{0.224} & \textbf{0.254} & \textbf{0.283}
& \textbf{0.143} & \textbf{0.236} & \textbf{0.269} \\

\ftmodel{Qwen3-Embed-0.6B†}
& 0.064 & 0.041 & 0.045
& 0.124 & 0.103 & 0.112
& 0.131 & 0.142 & 0.158
& 0.050 & 0.070 & 0.076 \\

Qwen3-Embed-0.6B
& 0.089 & 0.060 & 0.063
& 0.104 & 0.084 & 0.089
& 0.139 & 0.139 & 0.151
& 0.066 & 0.091 & 0.106 \\

Qwen3-Embed-4B
& 0.105 & 0.070 & 0.073
& 0.144 & 0.112 & 0.119
& 0.169 & 0.174 & 0.188
& 0.076 & 0.115 & 0.129 \\

\midrule

OpenAI (text-embedding-3-large)
& 0.092 & 0.065 & 0.069
& 0.140 & 0.110 & 0.116
& 0.176 & 0.181 & 0.195
& 0.084 & 0.121 & 0.136 \\

\bottomrule
\end{tabular}

\caption{Tier-2 ontology relation tasks evaluated using nDCG@k.}
\label{tab:tier2-ndcg}
\end{table*}

  \begin{table*}[t]
  \centering
  \scriptsize
  \begin{tabular}{l|ccc|ccc|ccc|ccc}
  \toprule
  & \multicolumn{12}{c}{MRR@k} \\
  \cmidrule(lr){2-13}
  & \multicolumn{3}{c|}{Dis$\rightarrow$Gene}
  & \multicolumn{3}{c|}{Dis$\rightarrow$Phen}
  & \multicolumn{3}{c|}{Phen$\rightarrow$Dis}
  & \multicolumn{3}{c}{Phen$\rightarrow$Gene} \\
  Model
  & @1 & @5 & @10
  & @1 & @5 & @10
  & @1 & @5 & @10
  & @1 & @5 & @10 \\
  \midrule

  BM25
  & 0.076 & 0.105 & 0.114
  & 0.084 & 0.120 & 0.128
  & 0.099 & 0.143 & 0.150
  & 0.048 & 0.060 & 0.062 \\

  MiniLM-L6-v2
  & 0.079 & 0.112 & 0.120
  & 0.120 & 0.175 & 0.187
  & 0.150 & 0.202 & 0.214
  & 0.058 & 0.084 & 0.090 \\

  \ftmodel{BioLORD-0.1B‡}
  & 0.112 & 0.164 & 0.177
  & 0.158 & 0.249 & 0.269
  & 0.034 & 0.125 & 0.150
  & 0.104 & 0.146 & 0.156 \\

  \ftmodel{BioLORD-0.1B†}
  & 0.061 & 0.090 & 0.099
  & 0.130 & 0.189 & 0.202
  & 0.151 & 0.219 & 0.232
  & 0.054 & 0.070 & 0.075 \\

  BioLORD-0.1B
  & 0.066 & 0.103 & 0.113
  & 0.099 & 0.152 & 0.163
  & 0.158 & 0.228 & 0.241
  & 0.060 & 0.084 & 0.088 \\

  MedEmbed-0.1B
  & 0.095 & 0.134 & 0.143
  & 0.148 & 0.211 & 0.225
  & 0.155 & 0.217 & 0.229
  & 0.079 & 0.106 & 0.113 \\

  SapBERT-0.1B
  & 0.096 & 0.131 & 0.141
  & 0.104 & 0.149 & 0.160
  & 0.142 & 0.206 & 0.217
  & 0.068 & 0.091 & 0.096 \\

  \midrule

  \ftmodel{Qwen3-Embed-0.6B‡}
  & \textbf{0.137} & \textbf{0.195} & \textbf{0.210}
  & \textbf{0.185} & \textbf{0.287} & \textbf{0.306}
  & \textbf{0.224} & \textbf{0.324} & \textbf{0.341}
  & \textbf{0.143} & \textbf{0.208} & \textbf{0.222} \\

  \ftmodel{Qwen3-Embed-0.6B†}
  & 0.064 & 0.089 & 0.097
  & 0.124 & 0.182 & 0.195
  & 0.131 & 0.193 & 0.206
  & 0.050 & 0.064 & 0.066 \\

  Qwen3-Embed-0.6B
  & 0.089 & 0.123 & 0.132
  & 0.104 & 0.149 & 0.158
  & 0.139 & 0.198 & 0.210
  & 0.066 & 0.083 & 0.089 \\

  Qwen3-Embed-4B
  & 0.105 & 0.142 & 0.152
  & 0.144 & 0.204 & 0.215
  & 0.169 & 0.238 & 0.251
  & 0.076 & 0.103 & 0.109 \\

  \midrule

  OpenAI (text-embedding-3-large)
  & 0.092 & 0.132 & 0.141
  & 0.140 & 0.198 & 0.209
  & 0.176 & 0.247 & 0.259
  & 0.084 & 0.110 & 0.116 \\

  \bottomrule
  \end{tabular}

  \caption{Tier-2 ontology relation tasks evaluated using MRR@k.}
  \end{table*}

\begin{table*}[t]
  \centering
  \scriptsize
  \begin{tabular}{l|ccc|ccc|ccc|ccc}
  \toprule
  & \multicolumn{12}{c}{Hit@k} \\
  \cmidrule(lr){2-13}

  & \multicolumn{3}{c|}{Dis$\rightarrow$Gene}
  & \multicolumn{3}{c|}{Dis$\rightarrow$Phen}
  & \multicolumn{3}{c|}{Phen$\rightarrow$Dis}
  & \multicolumn{3}{c}{Phen$\rightarrow$Gene} \\

  Model
  & @1 & @5 & @10
  & @1 & @5 & @10
  & @1 & @5 & @10
  & @1 & @5 & @10 \\
  \midrule

  BM25
  & 0.076 & 0.160 & 0.232
  & 0.084 & 0.184 & 0.245
  & 0.099 & 0.218 & 0.277
  & 0.048 & 0.081 & 0.100 \\

  MiniLM-L6-v2
  & 0.079 & 0.172 & 0.236
  & 0.120 & 0.273 & 0.369
  & 0.150 & 0.295 & 0.379
  & 0.058 & 0.128 & 0.173 \\

  \ftmodel{BioLORD-0.1B‡}
  & 0.112 & 0.258 & 0.349
  & 0.158 & 0.419 & 0.570
  & 0.034 & 0.342 & 0.524
  & 0.104 & 0.225 & 0.306 \\

  \ftmodel{BioLORD-0.1B†}
  & 0.061 & 0.145 & 0.213
  & 0.130 & 0.298 & 0.397
  & 0.151 & 0.341 & 0.435
  & 0.054 & 0.104 & 0.143 \\

  BioLORD-0.1B
  & 0.066 & 0.171 & 0.248
  & 0.099 & 0.248 & 0.334
  & 0.158 & 0.354 & 0.449
  & 0.060 & 0.124 & 0.159 \\

  MedEmbed-0.1B
  & 0.095 & 0.203 & 0.274
  & 0.148 & 0.326 & 0.429
  & 0.155 & 0.328 & 0.415
  & 0.079 & 0.155 & 0.204 \\

  SapBERT-0.1B
  & 0.096 & 0.197 & 0.272
  & 0.104 & 0.232 & 0.314
  & 0.142 & 0.319 & 0.401
  & 0.068 & 0.131 & 0.169 \\

  \midrule

  \ftmodel{Qwen3-Embed-0.6B‡}
  & \textbf{0.137} & \textbf{0.303} & \textbf{0.410}
  & \textbf{0.185} & \textbf{0.470} & \textbf{0.611}
  & \textbf{0.224} & \textbf{0.498} & \textbf{0.622}
  & \textbf{0.143} & \textbf{0.319} & \textbf{0.422} \\

  \ftmodel{Qwen3-Embed-0.6B†}
  & 0.064 & 0.140 & 0.202
  & 0.124 & 0.286 & 0.378
  & 0.131 & 0.302 & 0.398
  & 0.050 & 0.088 & 0.106 \\

  Qwen3-Embed-0.6B
  & 0.089 & 0.186 & 0.255
  & 0.104 & 0.228 & 0.298
  & 0.139 & 0.305 & 0.391
  & 0.066 & 0.117 & 0.163 \\

  Qwen3-Embed-4B
  & 0.105 & 0.211 & 0.288
  & 0.144 & 0.308 & 0.395
  & 0.169 & 0.360 & 0.456
  & 0.076 & 0.149 & 0.192 \\
\midrule
  OpenAI (text-embedding-3-large)
  & 0.092 & 0.204 & 0.277
  & 0.140 & 0.302 & 0.387
  & 0.176 & 0.372 & 0.466
  & 0.084 & 0.167 & 0.202 \\
  \bottomrule
  \end{tabular}

  \caption{Tier-2 ontology relation tasks evaluated using Hit@k.}
  \label{tab:tier2-hit}
  \end{table*}

  \begin{table*}[t]
  \centering
  \footnotesize
  \setlength{\tabcolsep}{3.5pt}
  \renewcommand{\arraystretch}{1.05}

  \begin{tabular}{l|ccc|ccc|ccc}
  \toprule

  & \multicolumn{9}{c}{PhenTriplet$\rightarrow$Dis} \\
  \cmidrule(lr){2-10}

  & \multicolumn{3}{c|}{nDCG@$k$}
  & \multicolumn{3}{c|}{MRR@$k$}
  & \multicolumn{3}{c}{Hit@$k$} \\

  Model
  & @1 & @5 & @10
  & @1 & @5 & @10
  & @1 & @5 & @10 \\

  \midrule

  BM25
  & 0.066 & 0.093 & 0.107
  & 0.066 & 0.103 & 0.110
  & 0.066 & 0.172 & 0.230 \\

  MiniLM-L6-v2
  & 0.072 & 0.106 & 0.123
  & 0.072 & 0.110 & 0.119
  & 0.072 & 0.180 & 0.246 \\

  \ftmodel{BioLORD-0.1B‡}
  & 0.085 & 0.196 & 0.244
  & 0.085 & 0.185 & 0.207
  & 0.085 & 0.368 & 0.527 \\

  \ftmodel{BioLORD-0.1B†}
  & 0.096 & 0.146 & 0.171
  & 0.096 & 0.151 & 0.163
  & 0.096 & 0.247 & 0.340 \\

  BioLORD-0.1B
  & 0.082 & 0.126 & 0.151
  & 0.082 & 0.131 & 0.143
  & 0.082 & 0.224 & 0.312 \\

  MedEmbed-0.1B
  & 0.083 & 0.122 & 0.143
  & 0.083 & 0.123 & 0.134
  & 0.083 & 0.202 & 0.281 \\

  SapBERT-0.1B
  & 0.076 & 0.111 & 0.131
  & 0.076 & 0.116 & 0.127
  & 0.076 & 0.194 & 0.272 \\

  \midrule

  \ftmodel{Qwen3-Embed-0.6B‡}
  & \textbf{0.183} & \textbf{0.275} & \textbf{0.313}
  & \textbf{0.183} & \textbf{0.281} & \textbf{0.297}
  & \textbf{0.183} & \textbf{0.449} & \textbf{0.572} \\

  \ftmodel{Qwen3-Embed-0.6B†}
  & 0.083 & 0.134 & 0.158
  & 0.083 & 0.135 & 0.147
  & 0.083 & 0.231 & 0.321 \\

  Qwen3-Embed-0.6B
  & 0.058 & 0.089 & 0.109
  & 0.058 & 0.092 & 0.102
  & 0.058 & 0.157 & 0.234 \\

  Qwen3-Embed-4B
  & 0.028 & 0.060 & 0.080
  & 0.028 & 0.061 & 0.071
  & 0.028 & 0.123 & 0.202 \\

  \midrule

  OpenAI (text-embedding-3-large)
  & 0.097 & 0.143 & 0.169
  & 0.097 & 0.149 & 0.160
  & 0.097 & 0.244 & 0.331 \\

  \bottomrule
  \end{tabular}

  \caption{
  Performance on the Tier-3
  PhenTriplet$\rightarrow$Dis retrieval task.
  We report nDCG, MRR, and Hit at
  $k \in \{1,5,10\}$.
  The † symbol identifies models fine-tuned using Tier~1
  supervision, while ‡ identifies models jointly fine-tuned
  using supervision from all three tiers.
  }
  \label{tab:tier3-combined}
  \end{table*}

 \clearpage

\onecolumn
\section{Error Analysis (Tier~3)}
\label{sup:error_analysis}

\paragraph{Failure-mode analysis.}
We analyzed Tier-3 queries for which the embedding model did not rank the
ground-truth disease first. For each failure, we computed phenotype overlap
between the query and the predicted disease
($\mathrm{Ov}_{\mathrm{pred}}$), the rank of the ground-truth disease under
the Phenomizer-style Resnik reference ($r_{\mathrm{Resnik}}$), and mean
query phenotype information content ($\overline{\mathrm{IC}}_q$).
Ontology recovery was defined as $r_{\mathrm{Resnik}}\leq10$; the
low-information threshold was $\overline{\mathrm{IC}}_q\leq5.949$, the
25th percentile among embedding failures.

Failures were assigned sequentially to mutually exclusive categories:
\textit{generic-bias} if
$\overline{\mathrm{IC}}_q\leq5.949$;
\textit{related-disease confusion} if
$\mathrm{Ov}_{\mathrm{pred}}=2$ and $r_{\mathrm{Resnik}}\leq10$;
\textit{aggregation failure} if
$\mathrm{Ov}_{\mathrm{pred}}\leq1$ and $r_{\mathrm{Resnik}}\leq10$; and
\textit{semantic drift} otherwise.

\paragraph{Interpretation.}
Tier-3 queries are phenotype triplets. Predictions matching two phenotypes
are treated as near misses; those matching at most one are aggregation
failures only when the ontology reference retrieves the true disease in
the top 10, and semantic drift otherwise. Low-information is an
operational query-specificity category, not a causal explanation.

\begin{table*}[!htbp]
\centering
\scriptsize
\setlength{\tabcolsep}{4pt}
\renewcommand{\arraystretch}{0.95}
\begin{tabularx}{\textwidth}{l l X X X c c c}
\hline
Mode & QID & Phenotypes & Predicted disease (embedding model) & True disease & $r_{\text{Phenomizer}}$ &
Ov$_{\text{pred}}$ & Ov$_{\text{true}}$ \\
\hline

Aggregation failure & Q2599 &
Sloping forehead; 2--4 toe cutaneous syndactyly; Chiari type II malformation &
Crossed polysyndactyly &
Lathosterolosis &
1 & 0 & 3 \\

Aggregation failure & Q4636 &
Finger syndactyly; Ungual fibroma; Glue ear &
Crossed polysyndactyly &
Choroidal atrophy-alopecia syndrome &
1 & 1 & 3 \\

Aggregation failure & Q5613 &
Long philtrum; Conical incisor; Hypophosphaturia &
Hypophosphatemic rickets, autosomal recessive, 2 &
Global developmental delay-osteopenia-ectodermal defect syndrome &
1 & 0 & 3 \\

Related disease confusion & Q4072 &
Flat face; Sacral dimple; Prominent fingertip pads &
PDE4D haploinsufficiency syndrome &
Vulto-van Silfout-de Vries syndrome &
1 & 2 & 3 \\

Related disease confusion & Q1943 &
Short philtrum; Prominent fingertip pads; Scoliosis &
1q21.1 microdeletion syndrome &
Global developmental delay-visual anomalies-progressive cerebellar atrophy-truncal hypotonia syndrome &
1 & 2 & 3 \\

Related disease confusion & Q2002 &
Midface retrusion; Scoliosis; Radial deviation of finger &
Autosomal dominant Robinow syndrome &
Acrofacial dysostosis 1, Nager type &
1 & 2 & 3 \\

Generic bias & Q18 &
Intellectual disability; Tremor; Tapered finger &
Developmental delay and seizures with or without movement abnormalities &
Agenesis of the corpus callosum with peripheral neuropathy &
1 & 2 & 3 \\

Generic bias & Q976 &
Abnormal cardiovascular system morphology; Long philtrum; Frontal bossing &
1q21.1 microdeletion syndrome &
Doors syndrome &
1 & 2 & 3 \\

Generic bias & Q730 &
Seizure; Splenomegaly; Ptosis &
Sialidosis type 2 &
Proteus syndrome, somatic &
1 & 2 & 3 \\

Semantic drift & Q6075 &
Ptosis; Paraplegia; Brain atrophy &
Primary lateral sclerosis &
Oculopharyngodistal myopathy 1 &
12 & 0 & 3 \\

Semantic drift & Q651 &
Brachydactyly; Limitation of joint mobility; Abnormal metaphysis morphology &
Metaphyseal dysostosis-intellectual disability-conductive deafness syndrome &
Thanatophoric dysplasia type 2 &
12 & 0 & 3 \\

Semantic drift & Q2375 &
Generalized-onset seizure; Lower limb spasticity; Status epilepticus &
Epileptic encephalopathy, early infantile, 6 (Dravet syndrome) &
Multiple congenital anomalies-hypotonia-seizures syndrome 2 &
13 & 1 & 3 \\

\hline
\end{tabularx}

\caption{
Representative Tier-3 embedding failures grouped by rule-based error mode. The embedding model frequently retrieves partially compatible disorders despite full phenotype coverage by the ground-truth disease, while the Phenomizer baseline often resolves the same queries through ontology-based phenotype aggregation. QID denotes the query identifier. $r_{\mathrm{Phenomizer}}$ denotes the ranking position assigned to the ground-truth disease by the ontology-aware Phenomizer baseline. 
Ov$_{\mathrm{pred}}$ and Ov$_{\mathrm{true}}$ denote phenotype overlap counts between the query phenotype set and the predicted or ground-truth disease, respectively.
}

\label{sup:table15}
\end{table*}



 \end{document}